\documentclass[runningheads]{llncs}
\usepackage[T1]{fontenc}
\usepackage{amssymb}
\usepackage{graphicx}
\usepackage{comment}
\usepackage{multirow}
\usepackage{cite} 
\usepackage{array}
\usepackage{hyperref}
\usepackage{tikz}
\usepackage{enumitem,kantlipsum}
\usepackage{amsmath}
\usepackage{makecell}
\usepackage[svgnames]{xcolor}
\usepackage{adjustbox}
\usepackage{tikz}

\usetikzlibrary{
	positioning,    
	shapes.geometric, 
	arrows.meta,    
	backgrounds     
}

\usepackage{booktabs}

\usepackage[ruled,vlined]{algorithm2e}
\usepackage{tikz}
\usetikzlibrary{arrows.meta, positioning, shapes.geometric}
\usetikzlibrary{shapes.geometric, arrows.meta, positioning}

\newcolumntype{P}[1]{>{\centering\arraybackslash}p{#1}}
\newcolumntype{C}[1]{>{\centering\let\newline\\\arraybackslash\hspace{0pt}}m{#1}}

\usepackage{pifont}
\usetikzlibrary{positioning}
\usepackage[most]{tcolorbox}
\tcbset{colback=gray!2,colframe=gray,boxsep=1mm,arc=1mm,left=1mm,right=1mm,top=0.5mm,bottom=0.5mm}
\usepackage{subcaption} 

\let\llncssubparagraph\subparagraph
\let\subparagraph\paragraph
\usepackage[compact]{titlesec}
\let\subparagraph\llncssubparagraph
\begin{document}
\title{Automated Skill Decomposition Meets Expert Ontologies: Bridging the Granularity Gap with LLMs \vspace{-0.5cm}}
\titlerunning{An Ontology-Grounded LLM Framework for Verifiable Skill Decomposition}
%
%

\author{LE Ngoc Luyen\inst{1,2}, Marie-Hélène ABEL\inst{1} \vspace{-0.2cm}
}
\institute{Université de technologie de Compiègne, CNRS, Heudiasyc (Heuristics and Diagnosis of Complex Systems), CS 60319 - 60203 Compiègne Cedex, France \and Gamaizer, 93340 Le Raincy, France \vspace{-0.5cm}}

\authorrunning{NL LE et al.}
%
%
\maketitle              
\begin{abstract}
This paper investigates automated skill decomposition using Large Language Models (LLMs) and proposes a rigorous, ontology-grounded evaluation framework. Our framework standardizes the pipeline from prompting and generation to normalization and alignment with ontology nodes. To evaluate outputs, we introduce two metrics: a semantic F1-score that uses optimal embedding-based matching to assess content accuracy, and a hierarchy-aware F1-score that credits structurally correct placements to assess granularity. We conduct experiments on \textit{ROME-ESCO-DecompSkill}, a curated subset of parents, comparing two prompting strategies: zero-shot and leakage-safe few-shot with exemplars. Across diverse LLMs, zero-shot offers a strong baseline, while few-shot consistently stabilizes phrasing and granularity and improves hierarchy-aware alignment. A latency analysis further shows that exemplar-guided prompts are competitive -- and sometimes faster -- than unguided zero-shot due to more schema-compliant completions. Together, the framework, benchmark, and metrics provide a reproducible foundation for developing ontology-faithful skill decomposition systems.
\end{abstract}
\vspace{-0.6cm}
\section{Introduction}

Accurately modeling skills at the right level of granularity underpins applications in personalized learning, job matching, curriculum design, and workforce upskilling~\cite{bloom2010taxonomy,mulder2017competence}. However, expert-created resources like the European Commission’s ESCO ontology~\cite{esco2022}, the U.S. O*NET database~\cite{onet}, or France’s Répertoire Opérationnel des Métiers et des Emplois (ROME) maintained by Pôle emploi~\cite{rome40}, while authoritative, often encode competencies at inconsistent depths and evolve too slowly to keep pace with technological change~\cite{futurereport}. This creates a critical granularity gap, where downstream systems require finer, more actionable decompositions of broad skills than these taxonomies provide~\cite{nationalskill,le2025vers}.

Large Language Models (LLMs) present a promising yet perilous solution. Recent work has shown that LLMs can assist in ontology population and taxonomy construction~\cite{lippolis2025ontology}. Given a broad skill, they can generate fine-grained sub-skills closer to teachable units or measurable outcomes. Yet, unguided generation is prone to hallucination, inconsistent specificity, and domain drift~\cite{huang2025survey}. Without principled grounding, such outputs can be difficult to verify against expert knowledge and thus risky to operationalize in high-stakes settings~\cite{bang2023multitask}.

This paper investigates how to harness the generative power of LLMs for skill decomposition while ensuring that outputs are verifiable and structurally sound. We propose a principled framework that treats an expert skill ontology not as a source of answers for the model, but as a gold-standard ruler for evaluation. We define the \emph{granularity gap} as the mismatch between high-level skill labels and the finer decomposition needed for adaptive learning paths or precise capability mapping. To control information leakage, we consider two prompting regimes: the ontology is withheld entirely (\emph{zero-shot})~\cite{kojima2022large}, or used only to derive illustrative, label-disjoint exemplars that steer style and specificity (\emph{few-shot})~\cite{brown2020language}.

To systematically test these approaches, we introduce an end-to-end pipeline that manages prompt construction, candidate generation, and normalization before aligning outputs to the ontology. Our evaluation moves beyond simple text matching by employing two key metrics: a semantic F1-score to measure content alignment~\cite{Reimers2019SentenceBERT} and a novel hierarchy-aware F1-score inspired by work in hierarchical classification~\cite{sun2001hierarchical}. We conduct our experiments on \textit{ROME-ESCO-DecompSkill}, a benchmark we curated from \textit{ESCO} and \textit{ROME}. To ensure comparability, we restrict evaluation to parent skills with a 5-12 direct children. Across models, zero-shot yields meaningful decompositions but tends to drift in depth. Few-shot, using label-disjoint exemplars, reduces this drift and stabilizes phrasing and granularity, particularly on mid-scale models. For very large models, the effect depends on exemplar choice: poorly matched exemplars can constrain coverage, revealing a precision–recall trade-off under stronger priors. Overall, ontology-derived structure functions as a useful prior, producing more reliable and taxonomically coherent skill decompositions.


The remainder of this paper is organized as follows. We begin in Section~\ref{section_relatedwork} by reviewing related work. Section~\ref{sec:proposition} details our methodology, from the formal task definition and our end-to-end pipeline to the specific decomposition methods. Building on this, Section~\ref{sec:experiment_settings} describes the experimental setup, and Section~\ref{sec:results} presents and analyzes the results. We conclude in Section~\ref{sec:conclusion}  with a summary of our findings.

\section{Related Work}\label{section_relatedwork}
The challenge of skill decomposition -- transforming a broad competence into a set of fine-grained sub-skills -- lies at the nexus of knowledge representation and generative artificial intelligence. Our work builds on established principles in ontology engineering but shifts the lens to evaluating the generative capabilities of LLMs. We first review the expert taxonomies that motivate the task, then examine prompting strategies that steer generation while controlling leakage and promoting appropriate granularity.

\subsection{Expert Taxonomies and Ontology-Based Modeling}
Authoritative resources such as \textit{ROME}, \textit{ESCO}, \textit{O*NET} enumerate competencies and relations for labor-market and training use cases \cite{esco2022,rome40,onet}. While they provide a crucial backbone, these resources often mix abstraction levels and evolve more slowly than practice \cite{di2023taxonomy}, which motivates automatic \emph{decomposition} of broad skills into actionable sub-skills. Ontology languages (OWL, RDF(S), SKOS) encode \texttt{hasSubSkill}/\texttt{narrower} hierarchies in machine-interpretable triples \cite{gruber1993ontology}; however, manual authoring at scale is costly. Our work bypasses this challenge by using these expert taxonomies not for construction, but for validation. We treat them as an external gold standard to evaluate generated decompositions and as a source for leakage-safe contextual evidence to regularize the model's output in terms of depth and terminology.

Prior research has extensively applied ontologies to structure and manage skills across various domains. Ontologies have been used to document competencies \cite{paquette2007ontology}, create personalized learning paths \cite{draganidis2006ontology}, and support continuous professional development within learning networks \cite{rezgui2014ontology}. More broadly, this work has led to the development of shared competency ontologies for the semantic web, designed to serve as common reference models for skill and profile descriptions \cite{paquette2021new}.

\subsection{LLM-Based Skill Decomposition}
LLMs (e.g., GPT-/5~\cite{gpt5}, LLaMA~\cite{touvron2023llama} families) can generate plausible sub-skill lists from high-level prompts and have been explored for curriculum design and competency mapping~\cite{li2023skillbert}. Research in prompt engineering has established techniques to shape and constrain outputs: \emph{zero-shot} prompting probes a model's latent knowledge without external guidance~\cite{kojima2022large,le2025well}, while \emph{few-shot} prompting uses exemplars to steer style and specificity~\cite{brown2020language}. Prior studies report that well-chosen exemplars can reduce variance in phrasing and help maintain the intended level of granularity.

In contrast to extraction or link prediction (which classify relations among given candidates), we study the \emph{skill decomposition} problem: producing a compact set of sub-skills from a single broad concept and evaluating them against an expert ontology using semantic and hierarchy-aware metrics. Our analysis quantifies how prompting strategies interact with ontology structure to improve semantic alignment and depth fidelity. The next section presents our methodology for ontology-grounded skill decomposition, including task formulation, pipeline design, and prompt templates for zero-shot and leakage-safe few-shot settings.

\section{Methodology}\label{sec:proposition}
In this section, we detail our approach for ontology-grounded evaluation of  LLM-based skill decomposition. We begin with a conventional task formulation that maps a broad skill to a compact set of fine-grained sub-skills and specifies feasibility constraints together with alignment to a reference ontology. We then outline the methodological framework -- prompt construction, candidate generation, normalization/deduplication, ontology alignment, and depth-aware scoring. Finally, we describe the decomposition methods instantiated within this framework: \emph{zero-shot} prompting and \emph{few-shot} prompting with curated, label-disjoint exemplars.

\subsection{Task Formulation}\label{sec:task}

Let the skill ontology be a labeled, typed, directed graph
$\mathcal{O}=(\mathcal{V},\mathcal{E},\mathcal{R},\ell)$, where $\mathcal{V}$ is the set of concepts,
$\mathcal{R}$ the relation types (e.g., \texttt{hasSubSkill}, \texttt{broader}, \texttt{narrower}),
$\mathcal{E}\subseteq \mathcal{V}\times \mathcal{R}\times \mathcal{V}$ the edge set, and
$\ell:\mathcal{V}\to\Sigma^{*}$ maps each node to its preferred label.
We denote by $\mathrm{Lex}(v)\subseteq \Sigma^{*}$ the lexical forms of $v$
taken from SKOS labels (\texttt{prefLabel}, \texttt{altLabel})~\cite{W3C-SKOS-Reference}.

Given a high-level (broad) skill $S_{\text{broad}}\in\mathcal{V}$, the generator receives
$(S_{\text{broad}},$ $ C, $ $ \mathcal{E}_{\mathrm{ctx}})$, where $C$ controls language, the target size $k$,
and decoding settings, and $\mathcal{E}_{\mathrm{ctx}}$ provides in-context information depending on
the prompting strategy: $\varnothing$ in zero-shot; a small curated set of exemplars in few-shot.
The model must produce an ordered list of $k$ fine-grained sub-skills
\[
\hat{\mathcal{G}}(S_{\text{broad}})=\big(s_1,\dots,s_k\big),\quad s_i\in\Sigma^{*},
\]
with $k$ typically in $[5,12]$ to keep decompositions actionable\footnote{Numbers 5 and 12 are illustrative defaults chosen for usability and labeling effort; any $k_{min}$, $k_{max}$	may be used.}.

The ontology provides the gold children
\[
\mathcal{G}(S_{\text{broad}})=\{\,g\in\mathcal{V}\mid (S_{\text{broad}},\texttt{\{hasSubSkill, narrower\}},g)\in\mathcal{E}\,\}.
\]

Generated outputs must satisfy: (i) \emph{relevance} (each $s_i$ refines $S_{\text{broad}}$);
(ii) \emph{strict specificity} ($s_i\neq \ell(S_{\text{broad}})$);
(iii) \emph{uniqueness} (no paraphrastic duplicates);
(iv) \emph{type conformity} (outputs are skills/competences, not tools or occupations).

\textit{Closed vs.\ open world}:
In the \emph{closed world},
\[
\mathcal{F}_{\mathrm{cw}}(S_{\text{broad}};d)=\{\, \ell(v)\mid v\in\mathcal{N}_d(S_{\text{broad}})\,\},
\]
predictions must come from an ontology neighborhood $\mathcal{N}_d(\cdot)$ (radius $d$ over
\texttt{hasSubSkill}/\texttt{narrower} edges). In the \emph{open world},
\[
\mathcal{F}_{\mathrm{ow}}(S_{\text{broad}})=\big(\Sigma^{*}\big)^{k},
\]
novel phrasings are allowed and later verified by alignment.

\textit{Alignment operator}:
Let $\hat{s}\in\hat{\mathcal{G}}(S_{\text{broad}})$ be a free-text predicted sub-skill. We align it to the ontology with
\[
\alpha:\ \Sigma^{*}\to \mathcal{V}\cup\{\bot\},\qquad
\alpha(\hat{s})=\arg\max_{v\in Desc(S_{\text{broad}})} \cos\!\big(E(\hat{s}),E(\ell(v))\big),
\]
where $Desc(S_{\text{broad}})$ is the set of all descendants of $S_{\text{broad}}$ (i.e., nodes reachable via \texttt{hasSubSkill}/\texttt{skos:narrower} edges), $\ell(\cdot)$ is the label map, and $E(\cdot)$ is a multilingual Sentence-BERT encoder~\cite{Reimers2019SentenceBERT}. We can accept the link $v^\star=\alpha(\hat{s})$ if $\cos\!\big(E(\hat{s}),E(\ell(v^\star))\big)\ge\tau$;
otherwise $\alpha(\hat{s})=\bot$.


\begin{figure}[t]
	\centering
	\includegraphics[width=0.65\textwidth]{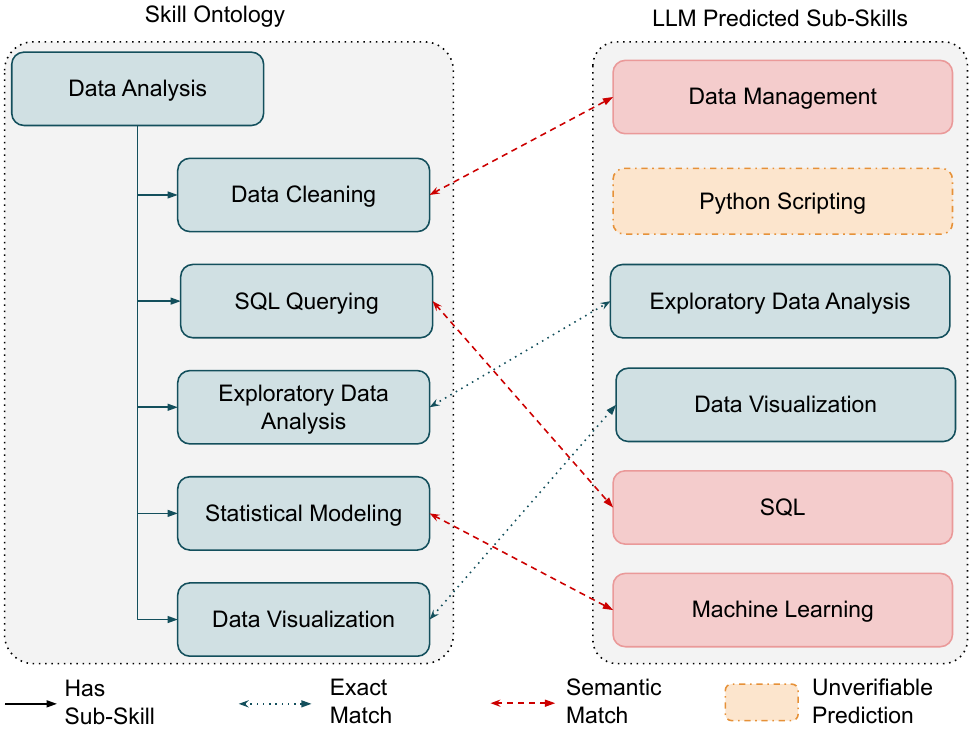}
	\caption{Example of ontology slice and LLM-based predictions with alignment results.}
	\label{fig:onto_pred_structure}
	\vspace{-0.6cm}
\end{figure}

Figure~\ref{fig:onto_pred_structure} illustrates the alignment step and the hierarchy-aware scoring on a toy example.
Let $S_{\text{broad}}=\textit{Data analysis}$ with gold children
$	\mathcal{G}(S_{\text{broad}})=\{
	\textit{data cleaning},\ $ $\textit{exploratory data analysis},\
	\textit{statistical modeling},$ $ 
	\textit{data visualization},\ $ $\textit{SQL querying}\}.$
A possible zero-shot list is
$
	\hat{\mathcal{G}}=\{
	\textit{exploratory data analysis},\  $ $\textit{python scripting},\
	\textit{data visualization},\ $ $ 
	\textit{machine learning modeling},\ $ $ \textit{SQL}\}.
$

We build a cosine-similarity matrix $S_{ij} $ $ = $ $ \cos(E(s_i), $ $ E(g_j))$ and apply maximum-weight $1$ -- $1$ matching.
In this example, \textit{exploratory data analysis} and \textit{data visualization} align to their gold counterparts;
\textit{SQL} aligns to \textit{SQL querying} (lexical equivalence);
\textit{machine learning modeling} may align to \textit{statistical modeling} if its similarity exceeds the acceptance
threshold; \textit{python scripting} remains unmatched when all similarities are low. These alignments quantify the
semantic fit between predictions and the gold set.

Each prediction $s_i$ is linked to its best concept among the descendants of $S_{\text{broad}}$. Exact children (e.g., \textit{exploratory data analysis}, \textit{data visualization},
and \textit{SQL} $\rightarrow$ \textit{SQL querying}) receive full credit; predictions that land strictly below a gold
child in the ontology (e.g., \textit{machine learning modeling} under \textit{statistical modeling}) receive partial
credit; items outside the subtree of any gold child (e.g., \textit{python scripting} if not placed under a gold child)
receive no credit. These credits modulate the semantic similarities ($H_{ij}=S_{ij}\times\text{credit}$), and the same
matching procedure on $H$ yields a hierarchy-aware assessment of the decomposition.

\subsection{Methodological Framework}\label{sec:overview}
\begin{figure}[t]
	\centering
	\includegraphics[width=0.95\textwidth]{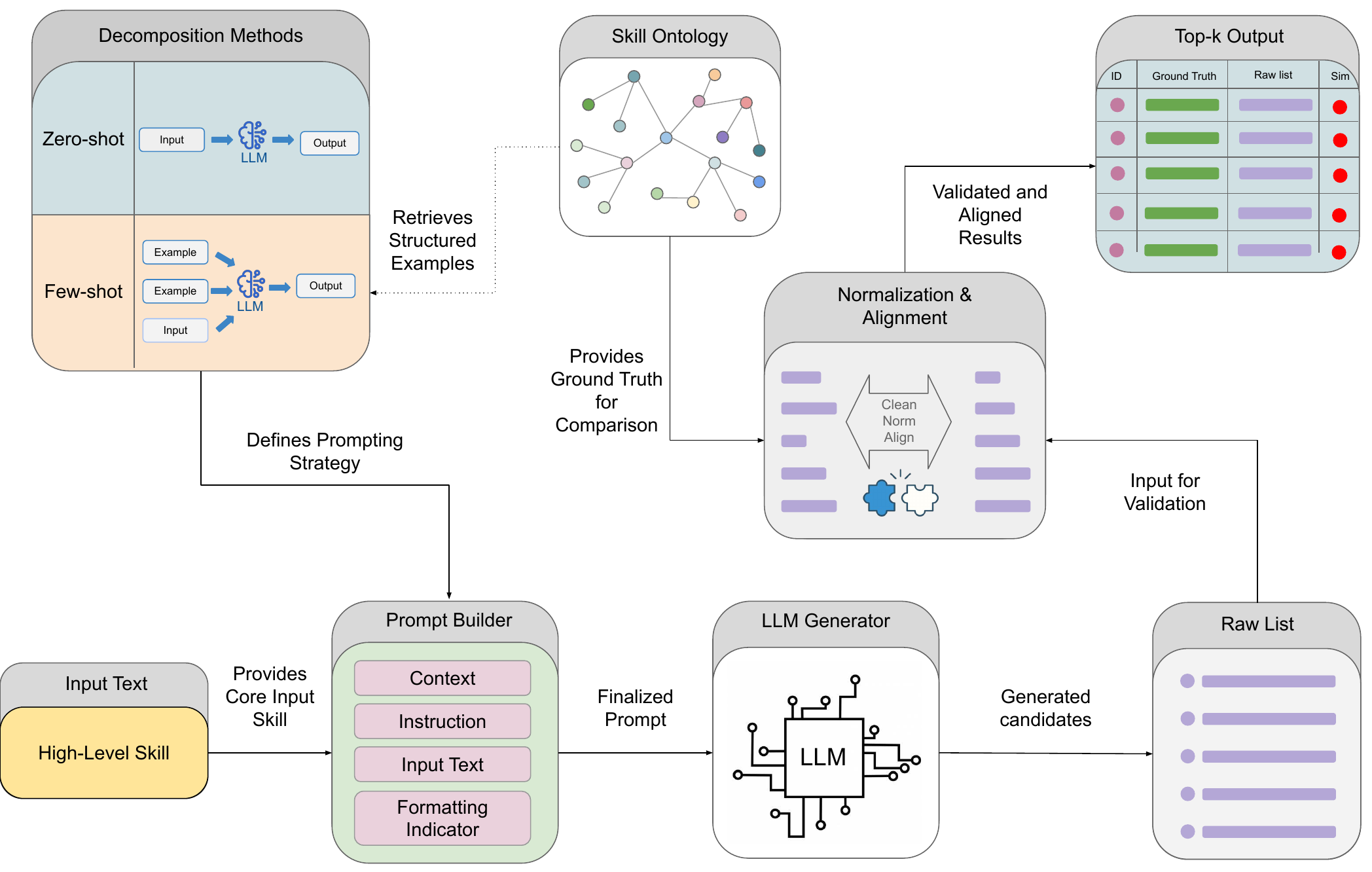}
	\caption{End-to-end framework for ontology-grounded skill decomposition: from $(S_{\text{broad}},C)$ the generator (optionally with exemplars) decodes exactly $k$ candidates, normalizes/deduplicates, aligns them to ontology nodes, and computes depth-aware scores. The ontology is used solely as external ground truth for alignment and evaluation.}
	\label{fig:block_pipeline}
	\vspace{-0.6cm}
\end{figure}

We develop an end-to-end framework that maps a broad input skill to a scored list of fine-grained sub-skills. Figure~\ref{fig:block_pipeline} instantiates the following blocks, which together govern prompt construction, candidate generation, post-processing, and ontology-grounded evaluation.

\textit{Input Text (High-Level Skill)}:
The input is a parent skill $S_{\text{broad}}$ together with controls $C$ (language, target size $k$, decoding settings). These define the task and expected output cardinality ($5\!\leq k\!\leq 12$).

\textit{Prompt Builder}:
This module converts $(S_{\text{broad}},C)$ into a structured instruction with explicit formatting constraints: noun-phrase only, exactly $k$ items, exclusion of tools/occupations, comma-separated output. The builder optionally inserts strategy-specific context $\mathcal{E}_{\mathrm{ctx}}$ while preserving the output schema to enable deterministic parsing.

\textit{Decomposition Methods}:
The framework supports two prompting strategies (detailed in Section~\ref{decomposition}): \emph{Zero-shot} (instruction only) and \emph{Few-shot} (instruction plus $k_{\mathrm{fs}}$ curated exemplars). These strategies differ only in how $\mathcal{E}_{\mathrm{ctx}}$ is populated by the Prompt Builder.

\textit{LLM Generator}:
Given the constructed prompt, the generator produces a raw ordered list of $k$ candidate sub-skills
$
\hat{\mathcal{G}}(S_{\text{broad}})= (s_1,\dots,s_k),
$
where each $s_i$ is a unique noun phrase; no item-level scores are used (the order reflects the model’s internal ranking).

\textit{Raw List}:
Immediately after decoding, we enforce lightweight syntactic and semantic checks before deeper processing:
(i) surface constraints (noun-phrase detector, length bounds);
(ii) parent repetition filter ($s_i\neq \ell(S_{\text{broad}})$);
(iii) coarse type check to exclude occupations/tools.

\textit{Normalization \& alignment:}
Candidates are lowercased, whitespace-normalized, and lightly cleaned
(punctuation/hyphenation). We suppress near-duplicates using a Sentence-BERT paraphrase
check (cosine on $E(\cdot)$); only one item per high-similarity cluster is kept.
We then apply the alignment function $\alpha$ defined in Section~\ref{sec:task}
(with the same candidate set $Desc(S_{\text{broad}})$ and threshold $\tau$).
Accepted items are either exact children ($v^\star\in\mathcal{G}(S_{\text{broad}})$)
or verifiable semantic matches ($v^\star\notin\mathcal{G}(S_{\text{broad}})$);
items with $\alpha(\hat{s})=\bot$ are marked unverifiable.

\textit{Top-$k$ output:}
The final artifact retains the $k$ generated items (in generation order) together with their alignments $\alpha(s_i)$; no item-level re-scoring is applied post hoc. Hierarchy awareness follows the evaluation code: each aligned item receives discrete credit based on its position in the subtree of $S_{\text{broad}}$ -- full credit for an exact child, partial credit for a deeper descendant of a gold child, and zero otherwise. This credit modulates semantic similarity for the hierarchy-aware assessment.

\textit{Skill ontology}: provides (i) the gold set $\mathcal{G}(S_{\text{broad}})$ used for alignment and scoring; and (ii) \emph{few-shot} exemplars $(S^{(j)},\mathcal{G}(S^{(j)}))$ chosen to be label-disjoint from $\mathcal{G}(S_{\text{broad}})$ to prevent leakage.

\subsection{Prompting Strategies for Skill Decomposition}\label{decomposition}

	\begin{figure}[!t]
		\centering
		\scriptsize
		\begin{subfigure}{.36\textwidth}
			\begin{tcolorbox}[title=Zero-shot - Prompt Template]
				\textbf{Context.} Parent skill $S_{\text{broad}}=[\text{SKILL\_NAME}]$; controls $C$ with $k=[k]$, language $=\mathrm{[LANG]}$; $\mathcal{E}_{\mathrm{ctx}}=\varnothing$.\\[2pt]
				\textbf{Instruction.} Decompose the broad skill $[\text{SKILL\_NAME}]$ into exactly $[k]$ fine-grained sub-skills. 
				Write each item as a short noun phrase, not a sentence; avoid verbs as heads. 
				Exclude tools, software names, and occupations; keep items directly relevant and one level more specific than the parent. 
				Avoid duplicates and near-paraphrases. \\[2pt]
				\textbf{Input Text.} Parent: $[\text{SKILL\_NAME}]$.\\[2pt]
				\textbf{Formatting Indicator.} Return exactly $k$ comma-separated items on one line: $s_1, s_2, \dots, s_k$.\\
				Language: $[\mathrm{LANG}]$.
			\end{tcolorbox}
			\vspace{-0.2cm}
			\caption{ZS}
			\label{fig:prompt_templates:zs}
		\end{subfigure}
		\begin{subfigure}{.56\textwidth}
			\begin{tcolorbox}[title=Few-shot — Prompt Template]
				\textbf{Context.} Parent $S_{\text{broad}}=[\text{SKILL\_NAME}]$; controls $C$ with $k=[k]$, language $=\mathrm{[LANG]}$;
				$\mathcal{E}_{\mathrm{ctx}}=\mathcal{E}_{\mathrm{fs}}=\{(S^{(j)},\mathcal{Y}^{(j)})\}_{j=1}^{k_{\mathrm{fs}}}$,
				with $S^{(j)}\neq S_{\text{broad}}$ and $\bigcup_j \mathcal{Y}^{(j)}\cap \mathcal{G}(S_{\text{broad}})=\emptyset$ (leakage-safe).\\[2pt]
				\textbf{Instruction.} Using the examples above as a guide for style and granularity, decompose $[\text{SKILL\_NAME}]$ into exactly $[k]$ sub-skills. 
				Match the exemplars’ specificity—neither broader nor substantially deeper—and use similar phrasing patterns. Do not copy labels verbatim from the examples. 
				\\[2pt]
				\textbf{Input Text.} \emph{Exemplars:}\\
				Parent $S^{(1)}$: [EX1\_PARENT] \quad Sub-skills $\mathcal{Y}^{(1)}$: [EX1\_S1], …, [EX1\_S5]\\
				Parent $S^{(2)}$: [EX2\_PARENT] \quad Sub-skills $\mathcal{Y}^{(2)}$: [EX2\_S1], …, [EX2\_S5]\\
				\emph{Target:} Parent $S_{\text{broad}}$: [SKILL\_NAME].\\[2pt]
				\textbf{Formatting Indicator.} Write short noun phrases; exclude tools, software names, and occupations; avoid duplicates/paraphrases: $s_1, s_2, \dots, s_k$. Language: $[\mathrm{LANG}]$.
			\end{tcolorbox}
			\vspace{-0.2cm}
			\caption{FS}
			\label{fig:prompt_templates:fs}
		\end{subfigure}
		\vspace{-0.3cm}
		\caption{Prompt templates by strategies (Context, Instruction, Input Text, Formatting Indicator).}
		
		\label{fig:prompt_templates}
		\vspace{-0.3cm}
	\end{figure}


We evaluate two prompting strategies that reuse the pipeline and notation from Section~\ref{sec:task} (Fig.~\ref{fig:block_pipeline}). To prevent lexical leakage, before prompting we ensure exemplars are label-disjoint with the target and remove any token-normalized forms of the target's gold children from the exemplar pool. Across strategies, only the \emph{in-context information} $\mathcal{E}_{\mathrm{ctx}}$ changes; the generator’s input/output schema remains as previously defined.

\subsubsection{Zero-shot (ZS)}
Instruction-only prompting with $\mathcal{E}_{\mathrm{ctx}}=\varnothing$.  
The instruction specifies the task (``decompose $S_{\text{broad}}$ into exactly $k$ fine-grained sub-skills'') and enforces the output schema (short noun phrases; no tools/occupations; comma-separated).  
This probes the model’s prior without external guidance; candidates then pass through the shared post-processing and alignment $\alpha$.

\subsubsection{Few-shot (FS)}
We augment the instruction with a small set of curated exemplars
\[
\mathcal{E}_{\mathrm{fs}}=\{(S^{(j)},\,\mathcal{Y}^{(j)})\}_{j=1}^{k_{\mathrm{fs}}}, \quad k_{\mathrm{fs}}\in\{2,3\},
\]
chosen from parents $S^{(j)}\neq S_{\text{broad}}$ such that $\mathcal{Y}^{(j)}\cap \mathcal{G}(S_{\text{broad}})=\emptyset$ (label-disjointness to avoid leakage).  
Exemplars are selected with a depth-proximity heuristic so that their children align with the target granularity $d^\star(S_{\text{broad}})$, guiding the phrasing and specificity without revealing gold labels.  
Decoding output is then normalized, de-duplicated, and aligned by the same $\alpha$ and depth analysis as ZS.

The prompt interface uses the templates shown in Figure~\ref{fig:prompt_templates}. All strategies share the same instruction, input text, and formatting requirements, differing only in their context sections:
\[
\mathcal{E}_{\mathrm{ctx}}=\begin{cases}
	\varnothing & \text{ZS (Fig.~\ref{fig:prompt_templates}\subref{fig:prompt_templates:zs})},\\
	\mathcal{E}_{\mathrm{fs}} & \text{FS (Fig.~\ref{fig:prompt_templates}\subref{fig:prompt_templates:fs})}.
\end{cases}
\]
Holding $\alpha$, $\mathcal{G}(S_{\text{broad}})$ and $d^\star$ constant isolates the causal effect of $\mathcal{E}_{\mathrm{ctx}}$ on the final top-$k$ output.

\section{Experimental Settings}\label{sec:experiment_settings}
We present the experimental setup, detailing the dataset, evaluated models, prompt configurations, and the metrics employed for performance assessment.
\subsection{Skill Ontology Dataset}
\label{sec:dataset}

Our experiments are grounded in the \textit{ROME-ESCO-DecompSkill} ontology, synthesised from \textit{ROME 4.0} and \textit{ESCO}. We focus on the \emph{Skills and Competences} pillar and use the French labels; the classification comprises 14,580 labelled concepts organised hierarchically. The core hierarchical property is \texttt{skos:broader}; we interpret its inverse, \texttt{skos:narrower}, as our target decomposition relation. Each concept carries lexical information via \texttt{skos:prefLabel} and \texttt{skos:altLabel}, which we use to align model predictions to ontology entries.

\begin{table}[t]
	\centering
	\caption{\textit{ROME-ESCO-DecompSkill} statistics. Parents are nodes with at least one \texttt{skos:narrower} child.}
	\label{tab:ontology_stats}
	\begin{tabular}{|C{6cm}|C{3cm}|}
		\hline
		\textbf{Statistic} & \textbf{Value} \\
		\hline
		Total labeled concepts (nodes) & 14580 \\\hline
		Parents with at least one child & 2857 \\\hline
		Total \texttt{skos:narrower} edges & 20822 \\\hline
		Average children per parent & 7.29 \\\hline
		Median children per parent & 2.00 \\\hline
		Std.\ dev.\ of children per parent & 18.34 \\\hline
		Min children per parent & 1 \\\hline
		Max children per parent & 270 \\\hline
		Parents with 5--12 children & 288 \\
		\hline
	\end{tabular}
	\vspace{-0.6cm}
\end{table}

Table~\ref{tab:ontology_stats} summarises the structural properties of the ontology. In total, 2,857 nodes act as parents with at least one child, linked by 20,822 \texttt{narrower} edges. The branching factor is highly skewed: the average number of children per parent is 7.29, while the median is only 2, with some transversal concepts reaching up to 270 children. For evaluation, we restrict to parents with between 5 and 12 direct children, yielding 288 eligible skills. From this pool we selected 288 broad skills ($S_{\text{broad}}$) to form the benchmark, ensuring coverage across diverse sectors.
Our implementation and datasets are publicly accessible\footnote{\href{https://github.com/lengocluyen/ROME-ESCO-DecompSkill}{https://github.com/lengocluyen/ROME-ESCO-DecompSkill}}.

\subsection{LLM Generators}

We evaluate seven LLM generators spanning diverse architectures and providers -- from open-weight to proprietary. All models are used off-the-shelf (no task-specific fine-tuning) under two prompting regimes: zero-shot (ZS) and leakage-safe few-shot (FS), as follows:

\begin{itemize}[topsep=0pt]
	\item \textit{OpenAI} -- \textit{GPT-5} (2025): proprietary transformer with advanced reasoning, multimodal I/O, and reinforcement learning from human feedback~\cite{gpt5}.
	\item \textit{OpenAI (open-weight)} -- \textit{GPT OSS 120B} (2025): Apache-2.0, open-weight model for transparent/local deployment~\cite{agarwal2025gpt}.
	\item \textit{DeepSeek AI} -- \textit{DeepSeek V3} (2024): multilingual transformer optimized for long context and code~\cite{liu2024deepseek}.
	\item \textit{Meta} -- \textit{LLaMA4 Scout} (2025): latency-optimized mixture-of-experts member of LLaMA-4~\cite{llama}.
	\item \textit{Kimi K2 AI} -- \textit{K2 Instruct} (2025): mid-scale instruction-tuned model aimed at efficiency and alignment~\cite{team2025kimi}.
	\item \textit{Mistral AI} -- \textit{Mistral Large} (2024): flagship model with efficient attention for fast inference~\cite{mistral}.
	\item \textit{Alibaba} -- \textit{Qwen3} (2025): large-scale transformer with extended context and multilingual alignment~\cite{yang2025qwen3}.
\end{itemize}

All models were accessed through official APIs or inference endpoints. We applied no task-specific fine-tuning, and evaluated them consistently under two prompting strategies: zero-shot and few-shot.

\subsection{Evaluation Metrics}\label{sec:metrics}

We evaluate decomposition quality per parent skill and report \emph{macro}-averages across all parents. Texts are embedded with a multilingual Sentence-BERT encoder $E(\cdot)$
(\texttt{paraphrase-multilingual-mpnet-base-v2}; mean pooling, L2 normalization) \cite{Reimers2019SentenceBERT}. Let $\hat{\mathcal{G}}_u=\{s_1,\dots,s_p\}$ be the model’s predictions for parent $u$ and $\mathcal{G}_u=\{g_1,\dots,g_q\}$ the gold children. We form a cosine-similarity matrix
\[
S_{ij}=\cos\!\big(E(s_i),E(g_j)\big), \qquad S\in\mathbb{R}^{p\times q}.
\]

\paragraph{Semantic Precision, Recall, and F1.}
We align predictions to gold items via maximum-weight $1$--$1$ matching using the Hungarian method \cite{Kuhn1955Hungarian,Munkres1957Assignment} on costs $C=1-S$.
Let $\Sigma_u=\sum_{(i,j)\in M_u} S_{ij}$ be the sum of matched similarities. Per-parent scores are
\[
P_{\text{sem}}^{(u)}=\frac{\Sigma_u}{p},\qquad
R_{\text{sem}}^{(u)}=\frac{\Sigma_u}{q},\qquad
F1_{\text{sem}}^{(u)}=\frac{2\,P_{\text{sem}}^{(u)}\,R_{\text{sem}}^{(u)}}{P_{\text{sem}}^{(u)}+R_{\text{sem}}^{(u)}+\varepsilon},
\]
with a small $\varepsilon$ for numerical stability; if $p{=}0$ or $q{=}0$ we set the scores to $0$. We then macro-average across parents:
\[
\overline{P}_{\text{sem}}=\tfrac{1}{|\mathcal{U}|}\sum_{u}P_{\text{sem}}^{(u)},\quad
\overline{R}_{\text{sem}}=\tfrac{1}{|\mathcal{U}|}\sum_{u}R_{\text{sem}}^{(u)},\quad
\overline{F1}_{\text{sem}}=\tfrac{1}{|\mathcal{U}|}\sum_{u}F1_{\text{sem}}^{(u)}.
\]
Because $F1$ is computed per parent and then averaged, $\overline{F1}_{\text{sem}}$ generally differs from $2\overline{P}_{\text{sem}}\overline{R}_{\text{sem}}/(\overline{P}_{\text{sem}}+\overline{R}_{\text{sem}})$.

\paragraph{Hierarchy-aware F1.}
To evaluate granularity, each prediction $s_i$ is \emph{linked} to its best concept among the descendants of $u$ (no depth cap). A prediction is accepted as linked if its cosine to the best concept is at least $\tau=0.78$. For any gold child $g_j$, we assign a discrete hierarchical credit:
\[
\text{credit}(s_i,g_j)=
\begin{cases}
	1.0, & \text{if the linked concept equals } g_j \text{ (exact child)},\\
	0.5, & \text{if the linked concept is a deeper descendant of } g_j,\\
	0,   & \text{otherwise.}
\end{cases}
\]
We then modulate semantic similarity with this credit,
$
H_{ij}=S_{ij}\cdot \text{credit}(s_i,g_j),
$
run the Hungarian method on $H$ (maximize total weight), and define per-parent scores as above (replace $S$ by $H$):
\[
P_{\text{hier}}^{(u)}=\frac{\Sigma^{\text{hier}}_u}{p},\quad
R_{\text{hier}}^{(u)}=\frac{\Sigma^{\text{hier}}_u}{q},\quad
F1_{\text{hier}}^{(u)}=\frac{2\,P_{\text{hier}}^{(u)}\,R_{\text{hier}}^{(u)}}{P_{\text{hier}}^{(u)}+R_{\text{hier}}^{(u)}+\varepsilon}.
\]
The reported \textbf{Hier-F1} is the macro-average $\overline{F1}_{\text{hier}}=\tfrac{1}{|\mathcal{U}|}\sum_{u}F1_{\text{hier}}^{(u)}$. This metric provides an assessment of each model's ability to decompose skills at appropriate levels of granularity while maintaining semantic accuracy, offering insights into both the precision of concept identification and the appropriateness of taxonomic positioning.

\section{Experimental Results}\label{sec:results}

We present the performance of the evaluated LLMs across prompting strategies, analyze their latency, and examine how they respond to ontology-derived exemplar signals; all results are averaged over the \textit{ROME-ESCO-DecompSkill} benchmark of parent skills.

\subsection{Performance Across Prompting Strategies}

Table~\ref{tab:llm_results} reports average semantic Precision, Recall, F1, and Hier-F1 across two prompting strategies. Zero-shot already provides a strong baseline, with most models landing around F1 $\approx 0.39$–$0.49$, indicating that large-scale pretraining equips LLMs with useful decomposition priors even without exemplars. Nonetheless, zero-shot outputs can drift in granularity or include overly broad items.

\begin{table*}[t]
	\centering
	\caption{Performance across LLMs and prompting strategies. Values are averaged over all parent skills.}
	\label{tab:llm_results}
	\begin{tabular}{|C{2.4cm}|C{1.8cm}|C{1.8cm}|C{1.8cm}|C{1.8cm}|C{1.8cm}|}
		\hline
		\textbf{LLM Generator} & \textbf{Strategy} & \textbf{Precision} & \textbf{Recall} & \textbf{F1} & \textbf{Hier-F1}\\
		\hline
		\multirow{2}{*}{DeepSeek V3}
		& Zero-shot & \underline{0.3844} & 0.5476 & \underline{0.4432} & 0.0656\\
		& Few-shot  & 0.3708 & \underline{0.5678} & 0.4400 & \underline{0.0879}\\
		\hline
		\multirow{2}{*}{GPT 5}
		& Zero-shot & 0.3729 & 0.5310 & 0.4299 & 0.0425\\
		& Few-shot  & \underline{0.3900} & \underline{0.5508} & \underline{0.4480} & \underline{0.0727}\\
		\hline
		\multirow{2}{*}{GPT OSS 120B}
		& Zero-shot & 0.4446 & \underline{0.5284} & \underline{0.4727} & \underline{0.0776}\\
		& Few-shot  & \underline{0.4692} & 0.4692 & 0.4302 & 0.0761\\
		\hline
		\multirow{2}{*}{K2 Instruct}
		& Zero-shot & 0.3916 & \underline{0.5095} & 0.4353 & 0.0387\\
		& Few-shot  & \underline{0.4402} & 0.5046 & \underline{0.4604} & \underline{0.0495}\\
		\hline
		\multirow{2}{*}{Llama4 Scout}
		& Zero-shot & \textbf{\underline{0.4488}} & 0.5583 & 0.4864 & 0.1220\\
		& Few-shot  & 0.4405 & \textbf{\underline{0.5817}} & \textbf{\underline{0.4902}} & \textbf{\underline{0.1399}}\\
		\hline
		\multirow{2}{*}{Mistral Large}
		& Zero-shot & 0.3880 & 0.5613 & 0.4503 & 0.0737\\
		& Few-shot  & \underline{0.3923} & \underline{0.5743} & \underline{0.4575} & \underline{0.0835}\\
		\hline
		\multirow{2}{*}{Qwen3}
		& Zero-shot & 0.3140 & 0.5382 & 0.3895 & \underline{0.0514} \\
		& Few-shot  & \underline{0.3146} & \underline{0.5383} & \underline{0.3900} & 0.0509\\
		\hline
	\end{tabular}
	\vspace{-0.5cm}
\end{table*}

Few-shot prompting generally stabilizes phrasing and depth, and it raises Hier-F1 for most models (e.g., \textit{DeepSeek V3}, \textit{GPT-5}, \textit{K2 Instruct}, \textit{Llama4 Scout}, \textit{Mistral Large}). The improvements in overall F1 are most noticeable for medium-scale models -- \textit{GPT-5} (0.4299 $\rightarrow$ 0.4480) and \textit{K2 Instruct} (0.4353 $\rightarrow$ 0.4604) -- while \textit{Mistral Large} and \textit{Llama4 Scout} see modest gains (0.4503 $\rightarrow$ 0.4575 and 0.4864 $\rightarrow$ 0.4902, respectively). \textit{Qwen3} changes are negligible, and \textit{GPT OSS 120B} shows lower F1 under few-shot due to a recall drop, despite higher precision. Overall, the best run in this comparison is \textit{Llama4 Scout} with few-shot (F1 = 0.4902, Hier-F1 = 0.1399), suggesting that exemplar guidance can improve both semantic fit and hierarchical placement when exemplars match the target depth.

Overall, ZS provides robust baselines but is prone to granularity drift. FS acts as a structural prior that tightens phrasing, improves precision, and enhances hierarchical placement -- especially for medium-scale models -- whereas very strong ZS systems may require carefully depth-aligned exemplar selection to avoid recall penalties.



\subsection{Latency Analysis}

\begin{figure}[t]
	\centering
	\includegraphics[width=0.98\linewidth]{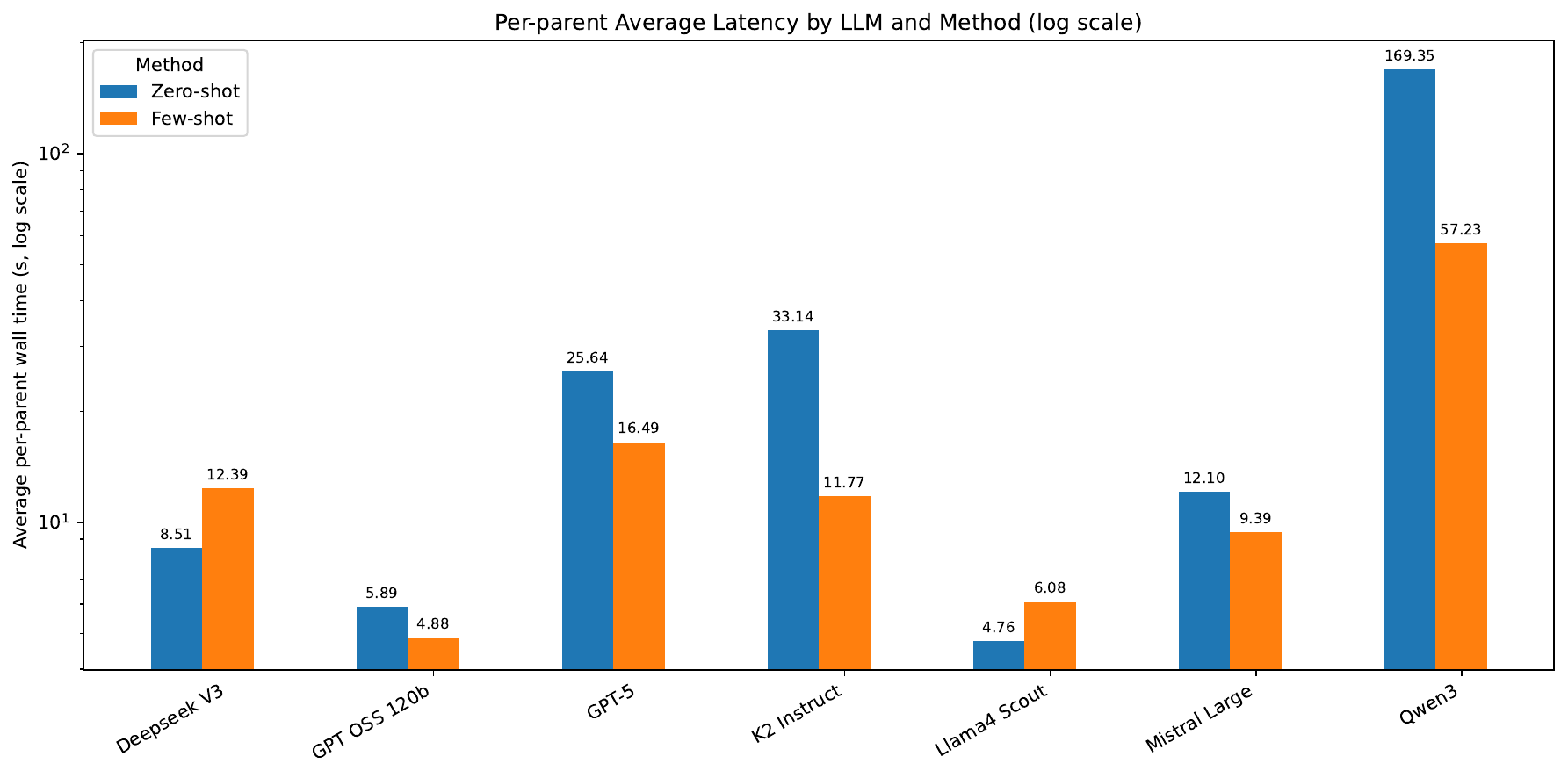}
	\vspace{-0.2cm}
	\caption{Per-parent average wall time (seconds, log scale) for Zero-shot and Few-shot prompting across LLMs.}
	\label{fig:latency_logscale}
	\vspace{-0.3cm}
\end{figure}

Figure~\ref{fig:latency_logscale} reports the per-parent average wall time (seconds, logarithmic scale) for the two prompting strategies. Latencies span over an order of magnitude -- from about 4.76\,s at the low end to 169.35\,s at the high end -- indicating substantial variability across model -- strategy combinations. Notably, few-shot is not uniformly faster or slower than zero-shot: in some cases average latency decreases (e.g., 8.51\,s $\rightarrow$ 5.89\,s), while in others it increases (e.g., 25.64\,s $\rightarrow$ 33.14\,s), reflecting the trade-off between longer prompts with exemplars versus potentially longer unconstrained completions under zero-shot.

Most configurations cluster in a mid-range (roughly 6 - 16\,s per parent), with a few pronounced outliers (e.g., $\sim$57.23\,s and $\sim$169.35\,s). These patterns suggest that prompt format and decoding behavior jointly determine wall time: exemplar-guided prompts can encourage concise, schema-compliant outputs that terminate earlier, whereas minimalist zero-shot instructions sometimes elicit longer continuations, offsetting their shorter input. Conversely, for some models the exemplar overhead dominates, yielding higher few-shot latency.

From an engineering perspective, these results emphasize that efficiency is model- and prompt-dependent. When deploying decomposition at scale, practitioners should benchmark both regimes with matched decoding limits, keep exemplars minimal and depth-aligned, and enforce strict output schemas (cardinality and surface form) to reduce variance in completion length without sacrificing quality.

\subsection{Association Between LLMs and Ontology-Derived Exemplar Signals}

ZS and FS reveal how LLMs respond to ontology-derived \emph{exemplar} signals (label-disjoint children from other parents). FS consistently reduces phrasing variance and suppresses over-general outputs, acting as a structural prior that steers generations toward the target depth and terminology. This is reflected in higher Hier-F1 for most models (e.g., \textit{DeepSeek V3}, \textit{GPT-5}, \textit{K2 Instruct}, \textit{Llama4 Scout}, \textit{Mistral Large}) and modest F1 gains, especially on medium-scale systems where inductive biases from pretraining are less dominant. In contrast, \textit{Qwen3} is essentially neutral under FS, and \textit{GPT OSS 120B} exhibits a precision–recall trade-off (precision up, recall down), indicating that exemplars can over-regularize very strong ZS models when their depth/style is not closely matched to the target.

These observations suggest that symbolic ontologies function as \emph{structural priors} even without direct retrieval: carefully chosen, depth-aligned exemplars regularize neural generation by mitigating hallucinations, improving specificity, and promoting consistent taxonomic placement. Practically, FS is most effective for medium-scale models, while for larger models exemplar sets should be kept minimal (2–3 items), label-disjoint, and tuned to the desired granularity and lexical register to avoid recall penalties.

\section{Conclusion}\label{sec:conclusion}
This paper systematically evaluated LLMs for skill decomposition using the \textit{ROME-ESCO-DecompSkill} benchmark. We developed an ontology-grounded evaluation framework combining semantic similarity with hierarchy-aware scoring to assess both lexical alignment and structural granularity. Focusing on zero-shot and leakage-safe few-shot prompting, we find that zero-shot offers a solid baseline, while few-shot generally stabilizes phrasing, improves specificity, and yields more consistent hierarchical placement. These results highlight the value of symbolic ontologies as structural priors that guide generative models toward appropriate granularity without relying on direct lookup.
Moving forward, we will investigate retrieval-augmented grounding with masked ontology evidence, extend to multilingual settings, explore adaptive exemplar selection and graph-constrained decoding, and assess downstream integrations in curriculum design and personalized learning.

\section*{Acknowledgments}
We warmly thank the Ikigai consortium led by the association Games for Citizens, the company Gamaizer, as well as the FORTEIM project (winner of the AMI CMA France 2030 call for projects), for their support and collaboration. Their contributions have provided significant added value to the completion of this research.

 \bibliographystyle{splncs04}
 \bibliography{references}
\end{document}